\documentclass[final,5p,times,twocolumn]{elsarticle}

\usepackage{graphicx}
\usepackage{booktabs}
\usepackage{amsmath,amssymb}
\usepackage{array}
\usepackage{tabularx}
\usepackage{ragged2e}
\usepackage{microtype}
\usepackage{placeins}
\usepackage{xurl}
\usepackage[hidelinks]{hyperref}
\usepackage{caption}
\usepackage{float}
\selectfont

\biboptions{sort&compress}

\begin{document}
\begin{frontmatter}

\title{When Participant Disjoint Validation Is Not Enough to Establish Generalization in Parkinson's Gait Classification}

\author[ru]{Md. Sifat}
\ead{mdsifat371@gmail.com}
\author[ru]{Sania Akter}
\ead{saniaaker@gmail.com}
\author[ru]{Akif Islam}
\ead{iamakifislam@gmail.com}
\author[ru]{Md. Ekramul Hamid}
\ead{ekram_hamid@ru.ac.bd}

\affiliation[ru]{organization={Department of Computer Science and Engineering, University of Rajshahi},
            addressline={Motihar},
            city={Rajshahi},
            postcode={6205},
            country={Bangladesh}}

\begin{abstract}
Participant-disjoint cross-validation prevents direct participant overlap,
but it may still produce optimistic estimates when acquisition site and
diagnosis are structurally dependent. We examined Parkinson's disease
classification using 159 pressure-walkway summary features from 133
WearGait-PD participants, including 65 participants with Parkinson's disease
and 68 controls. The cohort was non-factorial: two sites contained only one
diagnostic group, whereas VA-Seattle was the only site containing both
groups. In repeated participant-level nested cross-validation, site alone
achieved an area under the receiver operating characteristic curve (AUC) of
$0.955\pm0.003$, exceeding the gait-feature model
($0.882\pm0.014$). Under the evaluated logistic-regression pipelines, adding
gait features to site, age, and sex produced no measurable improvement in
pooled discrimination. More site-controlled evaluations produced lower and
imprecise estimates: VA-Seattle same-site AUC was 0.668
(95\% bootstrap confidence interval 0.477--0.836), and the
JHOC + Bayview-to-VA-Seattle transport stress test achieved 0.576
(0.375--0.770). Supporting within-diagnosis site-prediction and
site-conditioned permutation analyses showed that the gait representation
retained substantial acquisition-site structure. These findings indicate
that participant-disjoint pooled validation did not isolate disease-related
discrimination from site-associated variation. They encourage careful
examination of site--label dependence and, where possible, support for
pooled results through within-site or independent validation before broader
claims are made about generalization across institutions.
\end{abstract}


\end{frontmatter}

\section{Introduction}

Parkinson's disease (PD) is a progressive neurodegenerative disorder in
which gait and balance impairment contribute substantially to falls,
disability, and reduced independence \cite{poewe2017pd}. Compared with
controls, people with PD commonly exhibit differences in gait speed, stride
and step length, swing and double-support timing, postural control, and
stride-to-stride variability \cite{zanardi2021gait,dibiase2020gait}.
Instrumented gait systems, including pressure-sensitive walkways, provide
objective summaries of spatial, temporal, pressure, asymmetry, and
variability characteristics
\cite{sanders2024walkway,lynall2017pkmas,parati2022walkway}. These
measurements have therefore been investigated for PD classification and
mobility assessment \cite{rehman2019gaitml,nanayakkara2025cop}.

High internal classification performance, however, does not necessarily
show that a model has learned a transportable disease-related signal.
Clinical datasets may differ across institutions in recruitment pathways,
population characteristics, acquisition procedures, equipment, and data
processing, leading to dataset shift and reduced transportability
\cite{zech2018generalization,dockes2021shift}. When these differences are
associated with the target label, models may exploit site-related shortcuts
rather than the intended clinical construct \cite{geirhos2020shortcut}.
Participant-disjoint cross-validation prevents the same participant from
appearing in both the training and test sets, but it does not remove a
site--label relationship that is reproduced across folds.

Model-selection bias and biased cross-validation can produce optimistic
performance estimates \cite{cawley2010selection,varma2006bias}.
Preprocessing leakage and limited reproducibility may further weaken the
reliability of medical machine-learning evidence
\cite{kapoor2023leakage,mcdermott2021reproducibility}. Current reporting and
risk-of-bias guidance therefore emphasizes transparent model development,
appropriate validation, uncertainty reporting, and assessment of
applicability
\cite{collins2024tripod,moons2025probast,futureai2025}. Broader
methodological critiques similarly caution against treating internal
predictive performance as evidence of clinical generalizability
\cite{cabitza2026wrong}. Recent audits of Parkinson's voice classification and brain-model embeddings have also highlighted demographic or acquisition-related structure that can complicate interpretation of pooled discrimination \cite{shukla2026voice,tao2026batch}.

\begin{figure*}[!t]
\centering
\begin{minipage}[t]{0.485\textwidth}
  \centering
  \includegraphics[width=\linewidth]{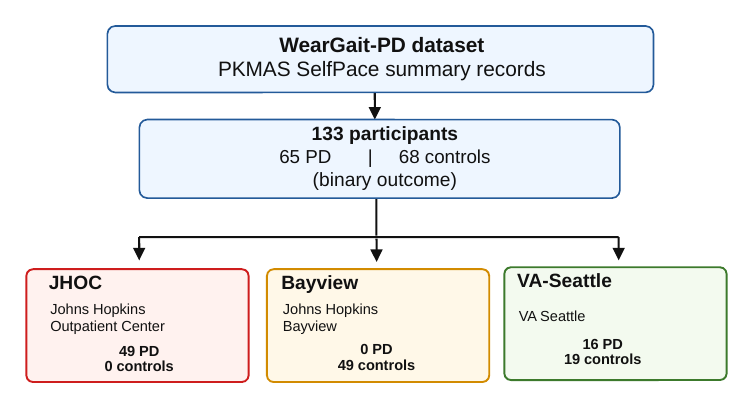}
  \captionof{figure}{Site--diagnosis composition of the primary SelfPace
  cohort. JHOC contained only participants with PD, Bayview contained only
  controls, and VA-Seattle was the only site containing both diagnostic
  groups.}
  \label{fig:cohort}
\end{minipage}\hfill
\begin{minipage}[t]{0.485\textwidth}
  \centering
  \includegraphics[width=\linewidth]{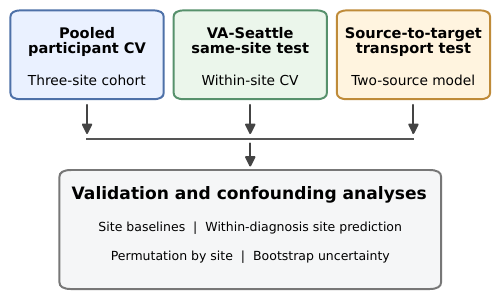}
  \captionof{figure}{Evaluation framework for pooled participant-disjoint,
  VA-Seattle same-site, and JHOC + Bayview-to-VA-Seattle transport analyses,
  supported by site-prediction, permutation, and uncertainty analyses.}
  \label{fig:evaluation-framework}
\end{minipage}
\end{figure*}

WearGait-PD is an open-access multimodal dataset containing inertial-sensor,
pressure-insole, synchronized-video, clinical, and ProtoKinetics
pressure-walkway data collected across three physical sites
\cite{anderson2026weargait,anderson2026dataset}. A separate WearGait-PD study assessed participant-disjoint Parkinson's classification using wearable inertial recordings, a different sensor modality and methodological focus from the present PKMAS audit \cite{herath2026multitask}. In the pressure-walkway
subset examined here, two sites contributed only one diagnostic group,
whereas VA-Seattle was the only site containing both participants with PD
and controls. This non-factorial structure allows acquisition site to act
as a surrogate for diagnosis in pooled analyses
(Fig.~\ref{fig:cohort}).

The original WearGait-PD publication is primarily a dataset descriptor and
does not report the specific pooled PKMAS diagnostic benchmark evaluated in
this study. The present work should therefore be understood as a
methodological investigation of a plausible downstream machine-learning use
of the released data, rather than as a reassessment of a clinical claim made
by the dataset creators.

We investigated whether participant-disjoint pooled validation could
separate disease-related gait discrimination from site-associated
structure. Specifically, we evaluated whether site alone reproduced pooled
classification performance, whether gait features added discrimination
beyond site and demographic variables, whether performance persisted within
the only mixed site and under source-to-target transport, and whether gait
features predicted acquisition site within diagnostic groups. By combining
non-clinical baselines, same-site evaluation, transport analysis,
within-diagnosis site prediction, and site-conditioned permutation, this
study illustrates why participant separation alone may be insufficient in
non-factorial multisite datasets.

\section{Materials and methods}

\subsection{Study design and analysis provenance}

This secondary analysis used previously collected, de-identified
WearGait-PD pressure-walkway data. The primary analysis included all
eligible participants with an exact \texttt{SelfPace} record; hurried-paced
walking was examined as a sensitivity condition. No prospective sample-size
calculation was performed because the cohort was fixed by the dataset
release.

Before the final rerun, the official data descriptor was used to correct
the physical-site mapping. An earlier exploratory grouping had merged JHOC
participants with PD and Bayview controls from distinct sites. Those results
were discarded, and all reported analyses use the official three-site
mapping.

Receiver operating characteristic AUC was the primary metric because no
clinical operating threshold or intended-use prevalence was defined
\cite{hanley1982auc}. Threshold-dependent metrics and decision-curve
analysis were not used to infer clinical utility because the study evaluated
validation performance rather than a deployable prediction model
\cite{vickers2006decision}.

\subsection{Dataset and site structure}

WearGait-PD includes 100 participants with physician-diagnosed PD and
85 controls \cite{anderson2026weargait}. Of these 185 participants, 133 had
both a PKMAS export and an exact \texttt{SelfPace} record and were included
in the analysis; the remaining 52 had no corresponding PKMAS export. The
analysed cohort comprised 65 participants with PD and 68 controls.

The official identifier mapping assigned NLS participants to JHOC, HC
participants to Bayview, and WPD/WHC participants to VA-Seattle. JHOC
contributed 49 participants with PD and no controls, Bayview contributed
49 controls and no participants with PD, and VA-Seattle contributed
16 participants with PD and 19 controls
(Fig.~\ref{fig:cohort}; Table~\ref{tab:cohort}).

Although the sites used common equipment models, software versions, and
experimental protocols \cite{anderson2026weargait}, site may still capture
technical, procedural, demographic, clinical, and other unmeasured
differences. We therefore interpret site as a composite marker of
acquisition context rather than a specific causal factor.

Site and diagnosis were strongly associated
($\chi^2=98.24$, $df=2$, $p=4.65\times10^{-22}$;
Cramer's $V=0.859$). Leave-one-site-out diagnosis validation was not feasible
because JHOC and Bayview each contained only one diagnostic class.

\begin{table}[H]
\caption{Self-paced cohort composition and basic demographics.}
\label{tab:cohort}
\centering
\footnotesize
\begin{tabular}{@{}llrrr@{}}
\toprule
Site & Group & $n$ & Age, years & Female \\
\midrule
JHOC & PD & 49 & $65.8\pm8.9$ (48) & 19/48 \\
Bayview & Control & 49 & $78.9\pm6.0$ (46) & 32/46 \\
VA-Seattle & PD & 16 & $70.3\pm5.8$ (16) & 4/16 \\
VA-Seattle & Control & 19 & $67.8\pm9.6$ (19) & 7/19 \\
\bottomrule
\end{tabular}
\begin{minipage}{\columnwidth}
\footnotesize Values are mean $\pm$ standard deviation (number available) or count/number available. Controls were participants without PD and were not required by the source protocol to be neurologically homogeneous.
\end{minipage}
\end{table}

\subsection{Feature preparation and model development}
The PKMAS export used a two-row header. Unique feature names were reconstructed by joining parameter and statistic labels, and task strings were stripped of whitespace. Only exact \texttt{SelfPace} records were retained; \texttt{SelfPace\_doorpat} represented a different doorway task. Participant suffixes such as \texttt{\_2} were collapsed to the base identifier, and the first deterministic participant--task record in source-row order was retained. Only two participants had more than one exact \texttt{SelfPace} row. Averaging all eligible rows for those participants produced materially unchanged results.

The source file contained 193 candidate parameters. Thirty-one columns containing ``\#Samples'' and three variables with more than 30\% missingness were removed, leaving 159 gait features. Missingness among retained features was minimal: 157 were complete, and the mean feature-level missing rate was below 0.15\% in every site--diagnosis group. No outcome-driven univariate feature selection preceded the multivariable analyses. Age, sex, height, and weight were extracted when available. Participants were included in each analysis only when the data required for that analysis were available; missing demographic information did not by itself exclude participants from analyses that did not require it. Additional clinical information available for participants with PD included disease duration, Hoehn and Yahr stage, total MDS-UPDRS-III, and deep-brain-stimulation status \cite{goetz2008updrs,hoehn1967stage}; the within-diagnosis site-prediction comparisons reported here used age and sex as measured covariates.

The primary estimator was L1-regularized logistic regression implemented with scikit-learn \cite{pedregosa2011sklearn}, using the \texttt{liblinear} solver, \texttt{max\_iter=5000}, no class weighting, and random state 42. Missing entries in retained continuous predictors were median-imputed, and continuous predictors were standardized, using training data only. Analyses requiring age or sex included only participants with the required demographic values available. Sex was encoded as female=1 and male=0, and site was represented by two indicators with Bayview as the reference. The regularization parameter was selected from $C\in\{0.03,0.1,0.3,1.0\}$ inside an inner stratified three-fold cross-validation loop. The outer evaluation used stratified five-fold cross-validation repeated 20 times. Repeat-level AUC was calculated from concatenated out-of-fold predictions for all eligible participants within each repeat. Every preprocessing, tuning, and fitting step was performed without access to the corresponding outer test participants.

Mean and standard deviation of repeat-level AUCs summarize variability under the specified split generator and are not population confidence intervals. Participant-bootstrap intervals were calculated separately when participant-level scores were available.

\subsection{Validation analyses}

Figure~\ref{fig:evaluation-framework} summarizes the evaluation framework.
Five pooled diagnosis models were evaluated: site only; age and sex; site,
age, and sex; all 159 gait features; and site, age, sex, and gait features.
Eligibility depended on the variables required by each model; paired
comparisons used the same eligible participants and outer partitions.
The site-only model served as a non-clinical baseline.

VA-Seattle was the only site containing both diagnostic groups. The repeated
nested protocol was rerun using its 35 participants. Each participant's
out-of-fold probability was averaged across the 20 repeats, and the
participant-level AUC was reported with a diagnosis-stratified percentile
interval from 5,000 bootstrap resamples. Repeat-level mean and standard
deviation were retained as a stability summary.

For the transport stress test, a source model was developed using JHOC
participants with PD and Bayview controls. Preprocessing and hyperparameter
selection used source data only, and the locked model was evaluated once at
VA-Seattle. Because diagnosis was perfectly confounded with source site,
this analysis was not treated as independent external validation. Target-site
uncertainty was estimated using 5,000 stratified participant-bootstrap
resamples.

To test whether gait encoded acquisition context while diagnosis was fixed,
site classifiers were evaluated separately among controls (Bayview versus
VA-Seattle, $n=68$) and participants with PD (JHOC versus VA-Seattle,
$n=65$). Each analysis compared measured covariates, gait features, and
covariates plus gait.

Finally, 200 global permutations shuffled diagnosis across all participants,
whereas 200 site-conditioned permutations shuffled labels separately within
site. These runs used the locked L1 logistic-regression pipeline with
$C=1.0$, stratified five-fold cross-validation, and fold-internal
preprocessing. Because JHOC contained only participants with PD and Bayview
only controls, 98 labels remained unchanged in every site-conditioned
permutation; only the 35 VA-Seattle labels could vary. The resulting AUC
distributions were interpreted as a structural sensitivity analysis rather
than a conventional exchangeability-based hypothesis test.

\subsection{Sensitivity analyses and reproducibility}
Sensitivity analyses evaluated missingness-only predictors, the 157 completely observed features, averaging duplicate records, joint stratification by the four observed site--diagnosis groups, ridge logistic regression, and the hurried-paced condition. The locked rerun used Python with NumPy, pandas, SciPy, scikit-learn, and Matplotlib. The analyses involved participant and site reconstruction, feature cleaning, nested cross-validation, bootstrap, permutation, and figure generation. Raw participant-level data must be obtained from Synapse under the original access terms.

\begin{table*}[!t]
\caption{Primary diagnosis-classification estimates.}
\label{tab:primary}
\centering
\footnotesize
\setlength{\tabcolsep}{4pt}
\begin{tabularx}{\textwidth}{@{}>{\RaggedRight\arraybackslash}p{0.29\textwidth}>{\RaggedRight\arraybackslash}X>{\RaggedLeft\arraybackslash}p{0.25\textwidth}@{}}
\toprule
Evaluation & Model input & AUC and uncertainty \\
\midrule
Pooled nested CV & Site only & $0.955\pm0.003$ \\
Pooled nested CV & Age and sex & $0.755\pm0.009$ \\
Pooled nested CV & Site, age, and sex & $0.953\pm0.005$ \\
Pooled nested CV & 159 gait features & $0.882\pm0.014$ \\
Pooled nested CV & Site, age, sex, and gait & $0.950\pm0.011$ \\
VA participant-level OOF AUC & 159 gait features & 0.668 (0.477--0.836) \\
JHOC + Bayview-to-VA-Seattle stress test & 159 gait features & 0.576 (0.375--0.770) \\
\bottomrule
\end{tabularx}
\begin{minipage}{0.94\textwidth}
\footnotesize Pooled results are mean $\pm$ standard deviation across 20 repeat-level AUCs. Values in parentheses are 95\% participant-bootstrap confidence intervals. OOF, out-of-fold.
\end{minipage}
\end{table*}

\section{Results}
\subsection{Site reproduced the pooled classification signal}
Acquisition site alone reproduced and exceeded the pooled gait-classification performance. Adding gait to site and demographics did not improve mean discrimination under the evaluated logistic-regression pipelines (Table~\ref{tab:primary}). In a paired fixed-regularization analysis using identical partitions, the AUC difference for adding gait was $-0.0002$ (bootstrap 95\% interval $-0.0020$ to 0.0011). This supports a narrow pipeline-specific conclusion rather than the absence of all possible incremental gait information.

\subsection{Site-controlled estimates were lower and imprecise}
Within VA-Seattle, participant-averaged out-of-fold predictions produced a same-site AUC of 0.668, but the wide bootstrap interval included chance and potentially meaningful discrimination. The small cohort therefore did not support a precise same-site performance claim. The JHOC + Bayview source model also produced a lower and imprecise estimate at VA-Seattle (Table~\ref{tab:primary}). Because diagnosis was perfectly confounded with source site, this analysis is interpreted as a transport stress test rather than external validation.

\begin{figure}[t]
\centering
\includegraphics[width=\columnwidth]{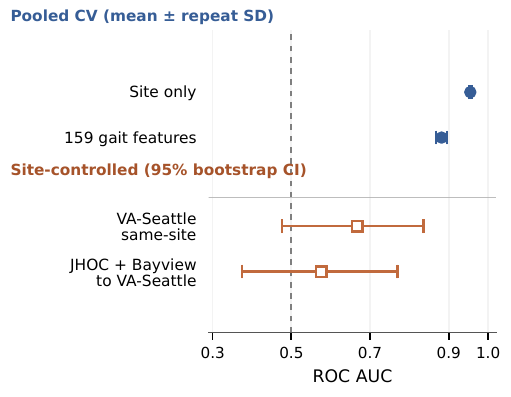}
\caption{Pooled and site-aware diagnosis-classification performance. Error
bars show repeat-level standard deviation or 95\% bootstrap confidence
intervals; the dashed line marks chance.}
\label{fig:pooled-vs-site-aware}
\end{figure}

\subsection{Gait features encoded acquisition-site structure}
Gait features predicted acquisition site within both diagnostic strata, with particularly strong discrimination among participants with PD. Measured covariates explained part of the between-site variation but did not remove the site-associated signal carried by the gait representation. Full estimates are reported in \ref{app:site}.

\subsection{Preserving site structure preserved apparent discrimination}
Global label permutation removed the site--diagnosis association and centred performance near chance. By contrast, site-conditioned permutation retained a high reference AUC distribution because 98 labels from the single-class sites remained structurally fixed and only the 35 VA-Seattle labels could vary. Preserving site-level class composition therefore retained most of the apparent discrimination (Fig.~\ref{fig:perm} and Table~\ref{tab:perm}).

\subsection{Sensitivity analyses}
Sensitivity analyses of missingness, complete features, duplicate averaging, joint stratification, ridge logistic regression, and hurried-paced walking reproduced the qualitative pattern. Full estimates are provided in \ref{app:sensitivity} and \ref{app:hurried}.

\section{Discussion}
\subsection{Principal finding}
Participant-disjoint pooled validation did not isolate disease-related generalization from acquisition-site structure. Site alone reproduced and exceeded the apparent gait-classification signal, adding gait did not improve discrimination beyond site and demographics, and performance was substantially weaker and more uncertain under same-site and source-to-target evaluation. Diagnosis-stratified analyses further showed that the gait representation itself encoded acquisition-site information, while the structural permutation demonstrated how the shortcut could persist.

\subsection{Why participant separation was insufficient}
Participant-disjoint cross-validation controls direct participant reuse, but it does not guarantee exchangeability with the intended target population. Every pooled fold contained JHOC participants with PD and Bayview controls. A classifier could therefore learn patterns that generalize to new participants from the same sites while remaining weakly related to PD pathology. This is structural shortcut learning rather than conventional duplicate-record leakage \cite{geirhos2020shortcut,kapoor2023leakage}. A related audit of Parkinson's voice screening identified demographic and site confounding in a different modality \cite{shukla2026voice}. The result extends observations from medical imaging to derived, interpretable tabular gait features and a linear model \cite{zech2018generalization}.

\subsection{Interpretation of site-associated structure}
The results do not establish that pressure-walkway features contain no PD-related information, that pressure-walkway assessment is clinically useless, or that WearGait-PD is invalid. Rather, the available non-factorial sampling design cannot separate disease-related gait information from site--diagnosis structure strongly enough to support the clinical interpretation implied by pooled AUC.

The WearGait-PD investigators used common equipment models, software, protocols, staff training, and quality-control procedures across sites \cite{anderson2026weargait}. Site should nevertheless be interpreted as a composite marker of technical, procedural, demographic, and clinical differences rather than as a single causal factor. Although measured covariates accounted for part of the observed variation, the non-factorial design and incomplete metadata prevented attribution of the remaining structure to any specific mechanism.

\subsection{Implications for medical machine learning}
Multisite prediction studies should reconstruct the site-by-label table before feature selection, evaluate site-only and demographic-only baselines on the same partitions as the clinical model, test whether clinical features add value beyond site and measured covariates, report within-site estimates wherever all target classes are represented, and reserve claims of clinical generalization for genuinely independent external validation. Pooled internal performance, within-site discrimination, source-to-target transport, and external validation answer different questions and should not be described interchangeably \cite{collins2024tripod,moons2025probast,futureai2025,cabitza2026wrong}.

\subsection{Strengths and limitations}
Strengths include use of the official physical-site mapping, disclosure of the corrected analysis provenance, matched participant partitions for paired analyses, training-fold-only preprocessing and nested model selection, a clearly defined participant-level same-site estimand, a separately labelled transport stress test, diagnosis-stratified site prediction, participant-level bootstrap uncertainty, structural permutation, and reproduction of the qualitative pattern in a second walking condition.

The only same-site cohort contained 35 VA-Seattle participants, producing wide uncertainty. JHOC and Bayview did not contain both diagnostic groups, preventing conventional multisite internal--external validation. Site and diagnosis were almost deterministically related, so pooled regression cannot identify independent causal effects of site and disease. Clinical covariates were incomplete, medication state was not standardized, and controls were not required to be neurologically homogeneous. Derived PKMAS summaries may obscure raw-signal mechanisms. The permutation analysis used a simplified fixed-parameter protocol and should be interpreted qualitatively. Finally, this is a case study of one dataset, and the proposed evaluation framework should be tested in other non-factorial multisite medical datasets.

\section{Conclusion}
In the analysed WearGait-PD pressure-walkway cohort, acquisition site and diagnosis were strongly entangled, allowing site alone to reproduce pooled PD classification performance. Same-site, transport, within-diagnosis site-prediction, and structural-permutation analyses showed that participant-disjoint pooled validation did not provide clear evidence of disease-related generalization. Multisite prediction studies should therefore examine site--label structure, include non-clinical baselines, report within-site performance, and reserve clinical generalization claims for genuinely independent validation.

\section*{Ethics statement}
The original WearGait-PD study obtained written informed consent, including consent for release of de-identified data. The original collection was approved under Johns Hopkins School of Medicine IRB00234370 with IRB Authorization Agreement 95-CDRH-2021-12-18 and VA IRB01702255 with IRB Authorization Agreement 2023-CDRH-095; the U.S. Food and Drug Administration was the relying institution \cite{anderson2026weargait}. The present study involved secondary analysis of repository-accessible, de-identified data, with no participant contact, intervention, or re-identification attempt.

\section*{Data and code availability}
WearGait-PD is available through Synapse accession \texttt{syn52540892} and DOI \href{https://doi.org/10.7303/syn52540892}{10.7303/syn52540892}, subject to repository access terms \cite{anderson2026dataset}. Analysis code is not publicly available at the time of this preprint. Raw participant-level data are not redistributed by the present authors.

\vspace{-7pt}
\section*{Funding}
This research did not receive any specific grant from funding agencies in the public, commercial, or not-for-profit sectors.

\vspace{-7pt}
\section*{Declaration of competing interest}
The authors declare that they have no known competing financial interests or personal relationships that could have appeared to influence the work reported in this paper.

\vspace{-7pt}
\section*{CRediT authorship contribution statement}
Md. Sifat: Conceptualization, Methodology, Software, Formal analysis, Investigation, Writing -- original draft, Writing -- review \& editing. Sania Akter: Data curation, Investigation, Validation, Writing -- review \& editing. Akif Islam: Validation, Resources, Writing -- review \& editing. Md. Ekramul Hamid: Supervision, Project administration, Writing -- review \& editing.

\vspace{-7pt}
\section*{Declaration of generative AI and AI-assisted technologies in the manuscript preparation process}
During the preparation of this work, the authors used OpenAI ChatGPT for language editing, document organization, and assistance in structuring reproducible analysis materials. After using this tool, the authors reviewed and edited the content, verified the references and numerical outputs, and take full responsibility for the content of the article.

\vspace{-7pt}
\section*{Acknowledgements}
The authors thank the WearGait-PD investigators for making the de-identified dataset and its documentation available to the research community.

\clearpage
\onecolumn
\appendix
\section{Cohort and feature-preparation details}
\label{app:cohort}
The complete clinical release contained 185 participants. Of these, 133 had at least one PKMAS export row; all 133 also had an exact \texttt{SelfPace} record. The remaining 52 participants had no PKMAS export row and were not eligible for the pressure-walkway analysis. Two participants had more than one exact \texttt{SelfPace} row. The primary analysis retained the first deterministic record in source-row order, while a sensitivity analysis averaged all eligible rows for those participants (Table~\ref{tab:flow}).

\begin{table}[H]
\centering\caption{Participant-flow accounting for the primary self-paced (SelfPace) analysis.}\label{tab:flow}
\small
\begin{tabularx}{0.74\textwidth}{@{}Xr@{}}
\toprule Stage & Count \\ \midrule
Complete clinical release & 185 \\
No PKMAS export row & 52 \\
At least one PKMAS export row & 133 \\
Exact \texttt{SelfPace} record & 133 \\
\texttt{SelfPace\_doorpat} excluded (additional rows) & 2 \\
Participants with repeated exact \texttt{SelfPace} records & 2 \\
Repeated rows beyond retained records & 2 \\
Final primary cohort & 133 \\
\bottomrule
\end{tabularx}
\end{table}

The PKMAS export contained 193 candidate parameters. Thirty-one internal columns containing ``\#Samples'' were removed because they describe the number of observations underlying a summary rather than gait behaviour. Three variables with more than 30\% missingness were also removed: Foot Angle Mean--ASI, DOP Mean--ASI, and FAP. The locked feature set therefore contained 159 gait variables. Of these, 157 were complete, and the mean feature-level missing rate was below 0.15\% in every site--diagnosis group. No outcome-driven univariate feature selection preceded the primary multivariable analyses (Table~\ref{tab:features}).

\begin{table}[H]
\centering\caption{Feature preparation in the primary self-paced analysis.}\label{tab:features}
\small
\begin{tabularx}{0.74\textwidth}{@{}Xr@{}}
\toprule Item & Count \\ \midrule
Candidate PKMAS parameters & 193 \\
Internal ``\#Samples'' columns removed & 31 \\
Variables removed for $>30\%$ missingness & 3 \\
Retained gait features & 159 \\
Completely observed retained features & 157 \\
Primary \texttt{SelfPace} participants & 133 \\
\bottomrule
\end{tabularx}
\end{table}

\section{Within-diagnosis acquisition-site prediction}
\label{app:site}
To assess whether the gait representation encoded acquisition context within a fixed diagnostic group, we predicted acquisition site separately among controls and participants with PD. The control analysis compared Bayview with VA-Seattle, whereas the PD analysis compared JHOC with VA-Seattle. Measured covariates comprised age and sex (Table~\ref{tab:sitepred}).

\begin{table}[H]
\centering\caption{Within-diagnosis acquisition-site prediction. Values are mean $\pm$ standard deviation from the corresponding participant-level validation analysis.}\label{tab:sitepred}
\small
\begin{tabularx}{0.94\textwidth}{@{}>{\RaggedRight\arraybackslash}Xccc@{}}
\toprule Diagnostic group (site comparison) & Age + sex & Gait & Age + sex + gait \\ \midrule
Controls (Bayview vs VA-Seattle) & $0.848\pm0.012$ & $0.883\pm0.041$ & $0.880\pm0.048$ \\
PD (JHOC vs VA-Seattle) & $0.749\pm0.041$ & $0.956\pm0.018$ & $0.956\pm0.018$ \\
\bottomrule
\end{tabularx}
\end{table}

\section{Robustness analyses}
\label{app:sensitivity}
Table~\ref{tab:robustness} summarizes analyses of missingness, duplicate-record handling, fold composition, and logistic-regression penalty. Missingness indicators alone performed at chance. Complete-feature, duplicate-averaged, and joint-stratified pipelines yielded pooled gait AUCs similar to the primary estimate. Ridge logistic regression produced the same qualitative conclusion, and a paired fixed-regularization comparison found no measurable AUC increase after gait features were added to site, age, and sex.

\begin{table}[H]
\centering\caption{Sensitivity analyses for the primary self-paced (SelfPace) cohort. Values support qualitative interpretation and are not independent confirmatory tests.}\label{tab:robustness}
\small
\begin{tabularx}{0.94\textwidth}{@{}>{\RaggedRight\arraybackslash}p{0.25\textwidth}>{\RaggedRight\arraybackslash}X>{\RaggedRight\arraybackslash}p{0.18\textwidth}@{}}
\toprule Analysis & Purpose & Result \\ \midrule
Missingness indicators only & Test whether measurement availability identified diagnosis & AUC 0.486 \\
157 complete features & Remove any contribution from feature imputation & AUC 0.895 \\
Average repeated eligible rows & Test deterministic first-record selection & AUC 0.893 \\
Joint site--diagnosis stratification & Preserve the four observed composition groups across folds & AUC 0.897 \\
Ridge logistic regression: gait & Test dependence on L1 regularization & AUC 0.838 \\
Ridge logistic regression: site, age, sex, and gait & Alternative combined estimator & AUC 0.909 \\
Paired addition of gait to site, age, and sex & Estimate incremental pooled discrimination on identical partitions & $\Delta$AUC $-0.0002$ $(-0.0020,0.0011)$ \\
\bottomrule
\end{tabularx}
\vspace{2pt}\begin{minipage}{0.96\textwidth}\footnotesize The interval for $\Delta$AUC is a participant-bootstrap 95\% interval.\end{minipage}
\end{table}

\section{Structural permutation analysis}
\label{app:permutation}
The permutation analysis used L1-regularized logistic regression with fixed $C=1.0$ and a newly generated shuffled stratified five-fold cross-validation split for each permutation; preprocessing remained internal to each training fold. We performed 200 permutations under each scheme. This simplified fixed-regularization protocol reduced computational cost and was not intended to reproduce the primary repeated nested-cross-validation point estimate. Its observed reference AUC was 0.908, compared with 0.882 under the primary repeated nested protocol.

\begin{figure}[H]
\centering
\includegraphics[width=0.84\textwidth]{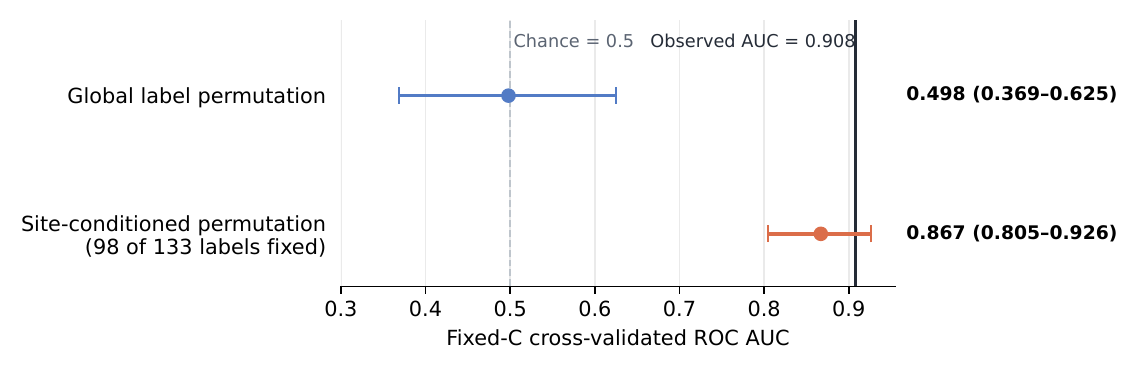}
\caption{Global and site-conditioned permutation reference distributions. Markers indicate the mean AUC across 200 permutations per scheme; horizontal intervals show the 2.5th--97.5th percentiles, and the solid vertical line marks the observed fixed-$C$ reference AUC of 0.908.}
\label{fig:perm}
\end{figure}

Global permutation broke the site--diagnosis association and produced a reference distribution centred near chance. Site-conditioned permutation preserved class counts within each physical site. Because JHOC contained only participants with PD and Bayview contained only controls, 98 of 133 labels remained unchanged in every site-conditioned permutation; only the 35 VA-Seattle labels could vary. We did not use a conventional empirical $p$-value for inference. Instead, we interpreted the distributions descriptively because the site-conditioned scheme did not satisfy conventional label exchangeability. The resulting reference AUC distribution remained high (Table~\ref{tab:perm} and Fig.~\ref{fig:perm}).

\begin{table}[H]
\centering\caption{Global and site-conditioned permutation reference summaries.}\label{tab:perm}
\footnotesize
\begin{tabularx}{0.77\textwidth}{@{}Xcc@{}}
\toprule Permutation scheme & Mean $\pm$ SD & 2.5th--97.5th \\ & & percentiles \\ \midrule
Global label permutation & $0.498\pm0.064$ & 0.369--0.625 \\
Site-conditioned permutation & $0.867\pm0.031$ & 0.805--0.926 \\
\bottomrule
\end{tabularx}
\end{table}

\section{Hurried-paced sensitivity analysis}
\label{app:hurried}
The hurried-paced condition contained 132 participant records after deterministic participant--task deduplication. We applied the same feature-cleaning and participant-level validation principles. The pooled AUC again exceeded the same-site and transport estimates, reproducing the qualitative pattern of the primary analysis (Table~\ref{tab:hurried}).

\begin{table}[H]
\centering\caption{Hurried-paced sensitivity results. Nested-cross-validation values are mean $\pm$ standard deviation across repeat-level AUCs. The value in parentheses is a 95\% participant-bootstrap confidence interval.}\label{tab:hurried}
\small
\begin{tabularx}{0.88\textwidth}{@{}>{\RaggedRight\arraybackslash}Xlc@{}}
\toprule Evaluation & Model input & AUC and uncertainty \\ \midrule
Pooled nested cross-validation & 159 gait features & $0.835\pm0.020$ \\
VA-Seattle same-site nested CV & 159 gait features & $0.506\pm0.072$ \\
JHOC + Bayview-to-VA-Seattle stress test & 159 gait features & 0.487 (0.296--0.684) \\
\bottomrule
\end{tabularx}
\end{table}

\section{Implementation and reproducibility}
\label{app:repro}
Figure~\ref{fig:workflow} summarizes the fold-internal feature and modelling workflow, and Table~\ref{tab:settings} lists the locked implementation settings. All continuous predictors were median-imputed and standardized using training-fold data only, and repeat-level AUC was calculated from concatenated out-of-fold predictions across all eligible participants within each repeat.

The locked rerun used Python 3 with NumPy 2.3.5, pandas 2.2.3, SciPy 1.17.0, scikit-learn 1.8.0, and Matplotlib 3.10.8. The analyses involved participant and site reconstruction, feature cleaning, nested cross-validation, bootstrap, permutation, aggregate output, and figure generation. Raw participant-level files are not redistributed and must be obtained from Synapse under the original repository access conditions.

\medskip
\noindent
\begin{minipage}[t]{0.50\textwidth}
\vspace{0pt}\centering
\includegraphics[width=0.97\linewidth]{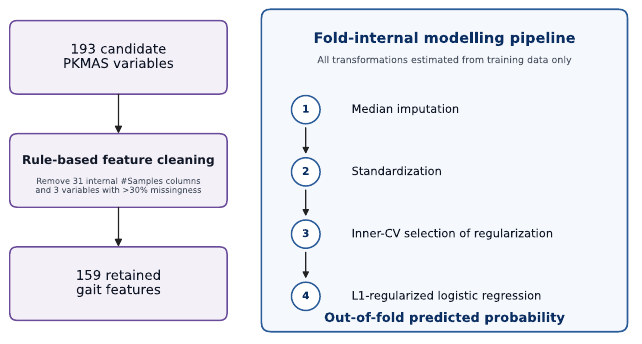}
\captionof{figure}{Feature preparation and fold-internal modelling workflow. Rule-based cleaning reduced 193 candidate variables to 159 retained gait features. Median imputation, standardization, inner-cross-validation selection of the regularization parameter, and model fitting were performed using training-fold data only.}
\label{fig:workflow}
\end{minipage}\hfill
\begin{minipage}[t]{0.47\textwidth}
\vspace{0pt}\centering
\captionof{table}{Locked implementation settings.}
\label{tab:settings}
\footnotesize
\setlength{\tabcolsep}{3pt}
\begin{tabularx}{\linewidth}{@{}>{\RaggedRight\arraybackslash}p{0.35\linewidth}>{\RaggedRight\arraybackslash}X@{}}
\toprule Component & Setting \\ \midrule
Primary estimator & L1 logistic regression \\
Solver & \texttt{liblinear} \\
Maximum iterations & 5000 \\
Class weighting & None \\
Random state & 42 \\
Candidate $C$ values & $\{0.03,0.1,0.3,1.0\}$ \\
Outer validation & Stratified five-fold CV, repeated 20 times \\
Inner model selection & Stratified three-fold CV \\
Participant bootstrap & 5,000 stratified resamples \\
Permutation analysis & Fixed $C=1.0$; one new shuffled stratified five-fold split per permutation; 200 permutations per scheme \\
\bottomrule
\end{tabularx}
\end{minipage}
\par\medskip

\FloatBarrier
\setlength{\bibsep}{1pt}
\small
\bibliographystyle{elsarticle-num-names}
\bibliography{references}

\end{document}